\pdfoutput=1

\documentclass[11pt]{article}

\usepackage{naacl2021}

\usepackage{times}
\usepackage{latexsym}

\usepackage[T1]{fontenc}

\usepackage[utf8]{inputenc}

\usepackage{microtype}

\usepackage{amsfonts}
\usepackage{bm}
\usepackage{array,amsmath}
\usepackage{amssymb}
\usepackage{dsfont}
\usepackage{float}
\usepackage{graphicx}
\usepackage{algorithm}
\usepackage{algorithmicx}
\usepackage{algpseudocode}
\usepackage{pgf,tikz}
\usepackage{mathrsfs}
\usepackage{multirow}
\usepackage{booktabs}
\usetikzlibrary{arrows}
\usepackage{threeparttable}
\usepackage[inline]{enumitem}

\def\figref#1{Figure~\ref{fig:#1}}
\def\figlabel#1{\label{fig:#1}\label{p:#1}}

\def\tabref#1{Table~\ref{tab:#1}}
\def\tablabel#1{\label{tab:#1}\label{p:#1}}

\def\eqref#1{Eq.~\ref{eqn:#1}}

\newcounter{notecounter}
\newcommand{\enotesoff}{\long\gdef\enote##1##2{}}
\newcommand{\enoteson}{\long\gdef\enote##1##2{{
			\stepcounter{notecounter}
			{\large\bf
				\hspace{1cm}\arabic{notecounter} $<<<$ ##1: ##2
				$>>>$\hspace{1cm}}}}}
\enoteson
\enotesoff
\long\def\eat#1{}

\title{Static Embeddings as Efficient Knowledge Bases?}

\author{Philipp Dufter\thanks{\mbox{\ \ } Equal contribution - random order.} , Nora Kassner$^{*}$, Hinrich Sch\"{u}tze\\
	Center for Information and Language Processing (CIS), LMU Munich, Germany\\
	{\tt \{philipp,kassner\}@cis.lmu.de}}

\begin{document}
\maketitle

\begin{abstract}

Recent research investigates factual knowledge stored in
large pretrained language models (PLMs). Instead of
structural knowledge base (KB) queries, masked sentences
such as ``Paris is the capital of [MASK]'' are used as
probes.  The good performance on this analysis task has been
interpreted as PLMs becoming potential repositories of
factual knowledge.  In experiments across ten linguistically
diverse languages, we study knowledge contained in static
embeddings. We show that, when restricting the output space
to a candidate set, simple nearest neighbor matching using
static embeddings performs better than PLMs. E.g., static
embeddings perform 1.6\% points better than BERT while just
using 0.3\% of energy for training.  One important factor in
their good comparative performance is that static embeddings
are standardly learned for a large vocabulary. In contrast,
BERT exploits its more sophisticated, but expensive ability
to compose meaningful representations from a much smaller
subword vocabulary.

\end{abstract}

\section{Introduction}

Pretrained language models (PLMs)
\cite{peters-etal-2018-deep,howard-ruder-2018-universal,devlin-etal-2019-bert}
can be finetuned to a variety of natural language processing
(NLP) tasks and then generally yield high
performance. Increasingly, these models and their generative
variants
(e.g., GPT, \citealp{brown2020language})
are used to solve tasks by simple text generation,
without any finetuning.
This motivated research on how much knowledge is
contained in PLMs: \citet{petroni2019language} used models pretrained
with a masked language objective to answer
cloze-style templates such as:
\begin{quote}
(Ex1) Paris is the capital of [MASK].
\end{quote}
Using this methodology, \citet{petroni2019language} showed that
PLMs capture some  knowledge implicitly. This
has been interpreted as suggesting that PLMs are promising as
repositories of factual knowledge. In this paper, we present
evidence that simple static embeddings like
fastText  
perform as well as
PLMs in the context of answering knowledge base (KB) queries. 
Answering KB queries can be decomposed into
two subproblems, \textbf{typing} and \textbf{ranking}. \textbf{Typing}
refers to the problem of predicting the correct type of the
answer entity; e.g., ``country'' is the correct type for
[MASK] in (Ex1), a task that PLMs seem to be good at. \textbf{Ranking} consists of 
finding the
entity of the correct type that is
the best fit (``France'' in (Ex1)).
By restricting the output space to the correct type we
disentangle the two subproblems and only evaluate
ranking. We do this for three reasons. (i) Ranking is the
knowledge-intensive step and thus the key research question.
(ii) Typed querying reduces PLMs' dependency on the
template. (iii) It allows a direct comparison between static word embeddings and PLMs. Prior work has adopted a similar approach \cite{Xiong2020Pretrained, Kassner2021MultilingualLI}.

\begin{table}
	\centering
	\footnotesize
	\def\mysep{0.2cm}
	\begin{tabular}{
			@{\hspace{\mysep}}l@{\hspace{\mysep}}|
			@{\hspace{\mysep}}r@{\hspace{\mysep}}|
			@{\hspace{\mysep}}r@{\hspace{\mysep}}
			@{\hspace{\mysep}}r@{\hspace{\mysep}}}
		\textbf{Model} & \textbf{Vocabulary Size} & \multicolumn{2}{c}{\textbf{p1}}\\
		&&LAMA&LAMA-UHN\\
		\midrule
		\midrule
Oracle & & 22.0 & 23.7 \\
\midrule BERT & 30k & 39.6 & 30.7 \\
mBERT & 110k & 36.3 & 27.4 \\
\midrule \multirow{7}{1cm}{fastText} & BERT-30k & 26.9 & 16.8 \\
& mBERT-110k & 27.5 & 17.8 \\
& 30k & 16.4 & 5.8 \\
& 120k & 34.3 & 25.0 \\
& 250k & 37.7 & 29.0 \\
& 500k & 39.9 & 31.8 \\
& 1000k & \textbf{41.2 }& \textbf{33.4 }\\
	\end{tabular}
\caption{Results for majority oracle, BERT, mBERT and
	fastText. Static fastText embeddings are competitive and
 outperform BERT for large vocabularies. 
BERT and mBERT use their subword vocabularies. For fastText,
	we
use
BERT/mBERT's vocabularies and newly trained wordpiece vocabularies on Wikipedia.
\tablabel{english}}
\end{table}

For a PLM like BERT, ranking amounts to finding the entity
whose embedding is most similar to the output embedding for
[MASK]. 
For static
embeddings, we rank entities (e.g., entities of type
country) with respect to similarity to the query entity
(e.g., ``Paris'' in (Ex1)). 
In experiments across ten
linguistically diverse languages, we show that this simple
nearest neighbor matching with fastText embeddings
performs comparably to or even better than BERT.
For example for English, fastText embeddings perform 1.6\% points
better than BERT (41.2\% vs.\ 39.6\%, see \tabref{english},
column ``LAMA'').
This suggests that
BERT's core mechanism  for answering factual queries is not more effective than simple nearest neighbor matching using fastText embeddings. 

We believe this means that
claims that PLMs are KBs have to be treated
with caution. Advantages of BERT are that it 
composes meaningful representations from a small subword
vocabulary and  handles typing
implicitly \cite{petroni2019language}. In contrast,
answering queries without restricting the answer space to a
list of candidates is hard to achieve with static word
embeddings. On the other hand, static embeddings
are cheap to obtain, even for large vocabulary sizes. This
has important implications for green NLP. PLMs require
tremendous computational resources, whereas static
embeddings have only 0.3\% of the carbon footprint of BERT
(see \tabref{emissions}). This argues for proponents of resource-hungry deep learning models
to try harder  to find cheap ``green''
baselines or to combine the best of both worlds (cf.\ \citealp{poerner-etal-2020-e}).

In summary, our contributions are:
\begin{enumerate}[label={\textbf{\roman{*})}}]
	\itemsep0em 
	\item We propose an experimental setup that allows a
	 direct comparison between  PLMs and static word
	 embeddings. We find that static word embeddings
	 show  performance similar to
	 BERT on the modified LAMA analysis task across ten languages.
	 \item We provide evidence that there is a trade-off between composing meaningful representations from subwords and increasing the vocabulary size. Storing information through composition in a network seems to be more expensive and challenging than simply increasing the number of atomic representations. 
        \item Our findings may point to a
	general problem: baselines that are
	simpler and ``greener''
are not given enough attention in deep learning.
\end{enumerate}
 Code and embeddings are available online.\footnote{\url{https://github.com/pdufter/staticlama}}

\section{Data}

We follow the LAMA setup introduced by \citet{petroni2019language}. More specifically, we use data from TREx \cite{elsahar-etal-2018-rex}. TREx consists of triples of the form (object, relation, subject). The underlying idea of LAMA is to query knowledge from  PLMs using templates without any finetuning: the triple (Paris, capital-of, France) is queried with the template ``Paris is the capital of [MASK].'' TREx has covers 41 relations. Templates for each relation were manually created by \citet{petroni2019language}. LAMA has been found to contain many ``easy-to-guess''
triples; e.g., it is easy to guess that a person with
an Italian sounding name is Italian. LAMA-UHN is a
subset of triples
that are ``hard-to-guess'' created by \newcite{poerner-etal-2020-e}.

Beyond English, we run experiments on nine additional languages
using mLAMA, a multilingual version of TREx \cite{Kassner2021MultilingualLI}. For an
overview of languages and language families
see \tabref{langs}.
For training static embeddings, we use Wikipedia dumps from October 2020.

\begin{table}
	\centering
	\footnotesize
	\begin{tabular}{llll}
		\textbf{Language} & \textbf{Code} & \textbf{Family} & \textbf{Script}\\
		\midrule
		\midrule
		Arabic & AR & Afro-Asiatic & Arabic\\
		German & DE & Indo-European & Latin\\
		English & EN & Indo-European & Latin\\
		Spanish & ES & Indo-European & Latin\\
		Finnish & FI & Uralic & Latin\\
		Hebrew & HE & Afro-Asiatic  & Hebrew \\
		Japanese & JA & Japonic & Japanese\\
		Korean & KO & Koreanic & Korean\\
		Turkish & TR & Turkic & Latin\\
		Thai & TH & Tai-Kadai& Thai
	\end{tabular}
\caption{Overview of the ten  languages in our experiments,
		including language family and script.\tablabel{langs}}
\end{table}
\section{Methods}

We describe our proposed setup, which allows to  compare  PLMs with static embeddings.

\subsection{PLMs}

We use the
following two  PLMs:
(i) BERT for English (BERT-base-cased, 
\citet{devlin-etal-2019-bert}), (ii)
mBERT for all ten languages (the multilingual
version BERT-base-multilingual-cased).

\citet{petroni2019language} use templates like ``Paris is the capital of [MASK]'' and give
$\arg\max_{w \in \mathcal{V}}p(w|t)$ as answer where $\mathcal{V}$ is
the vocabulary of the PLM and $p(w|t)$ is the probability
that word $w$ gets predicted in the template $t$. 

We follow the same setup as \cite{Kassner2021MultilingualLI} and use typed querying: for each relation, we create a
candidate set $\mathcal{C}$ and then predict
$\arg\max_{c\in \mathcal{C}}p(c|t)$. For most templates,
there is only one valid entity type, e.g., country for (Ex1).
We choose as $\mathcal{C}$ the
set of objects across all triples for a single relation.
The candidate set could also be obtained from an entity
typing system
(e.g., \citealp{yaghoobzadeh2018corpus}), but this
is beyond the scope of this paper.  Variants of typed
prediction have been used before \cite{Xiong2020Pretrained}.

We accommodate
multi-token objects, i.e., objects that are not contained in the vocabulary, by including multiple [MASK] tokens in
the templates. We then compute an object's score 
as the average of the log probabilities for
its individual tokens. Note that we do not perform any finetuning.

\begin{table}
	\centering
	\scriptsize
	\def\mysep{0.08cm}
	\begin{tabular}{
			@{\hspace{\mysep}}l@{\hspace{\mysep}}
			@{\hspace{\mysep}}r@{\hspace{\mysep}}
			@{\hspace{\mysep}}r@{\hspace{\mysep}}
			@{\hspace{\mysep}}r@{\hspace{\mysep}}
			@{\hspace{\mysep}}r@{\hspace{\mysep}}
			@{\hspace{\mysep}}r@{\hspace{\mysep}}
			@{\hspace{\mysep}}r@{\hspace{\mysep}}
			@{\hspace{\mysep}}r@{\hspace{\mysep}}
			@{\hspace{\mysep}}r@{\hspace{\mysep}}
			@{\hspace{\mysep}}r@{\hspace{\mysep}}
			@{\hspace{\mysep}}r@{\hspace{\mysep}}}
		 & \textbf{Vocab.}  & \multicolumn{9}{c}{\textbf{p1}}\\
		\textbf{Model} & \textbf{Size} & AR & DE & ES & FI & HE & JA & KO & TH & TR\\
		\midrule
		\midrule
Oracle & & 21.9 & 22.3 & 21.6 & 21.3 & 22.9 & 21.3 & 21.7 & 23.7 & 23.5 \\
\midrule mBERT & 110k & 17.2 & 31.5 & 33.6 & 20.6 & 17.5 & 15.1 & 18.9 & 13.5 & 33.8 \\
\midrule \multirow{6}{1cm}{fastText} 
& mB-110k & 16.4 & 20.9 & 24.6 & 21.4 & 14.5 & 12.9 & 16.1 & 12.9 & 26.0 \\
& 30k & 20.8 & 16.2 & 17.1 & 16.7 & 21.4 & 14.6 & 17.3 & 21.3 & 22.1 \\
& 120k & 27.9 & 25.2 & 31.0 & 24.2 & 28.3 & 22.4 & 28.2 & 28.0 & 33.2 \\
& 250k & 30.1 & 30.3 & 34.2 & 28.8 & 32.8 & 24.9 & 30.5 & 31.6 & 35.6 \\
& 500k & \textbf{31.7 }& 32.5 & \textbf{36.6 }& 30.9 & 33.7 & 27.0 & \textbf{31.5 }& \textbf{31.8 }& 36.1 \\
& 1000k & 31.3 & \textbf{33.6 }& 36.5 & \textbf{31.8 }& \textbf{33.9 }& \textbf{27.2 }& 29.8 & 30.5 & \textbf{36.6 }\\
	\end{tabular}
	\caption{p1 for mBERT and
	fastText on mLAMA. fastText
	clearly outperforms mBERT for large
        vocabularies. Numbers across languages are not
        comparable as the number of triples varies. 
\tablabel{multilingual}}
\end{table}

\begin{table}
	\centering
	\footnotesize
	\def\mysep{0.1cm}
	\begin{tabular}{
			@{\hspace{\mysep}}l@{\hspace{\mysep}}
			@{\hspace{\mysep}}r@{\hspace{\mysep}}
			@{\hspace{\mysep}}r@{\hspace{\mysep}}
			@{\hspace{\mysep}}r@{\hspace{\mysep}}
			@{\hspace{\mysep}}r@{\hspace{\mysep}}}
		\textbf{Model} & \textbf{Power (W)} & \textbf{h} & \textbf{kWh $\cdot$ PUE} & \textbf{CO\textsubscript{2}e} \\
		\midrule
		\midrule
		BERT & 12,041 & 79 & 1,507 & 1,438 \\
		fastText-en & 618 & 5 & 5 & 5\\
		\midrule
		ratio-en & 0.05 & 0.06 & 0.003 & 0.003
	\end{tabular}
	\caption{Power consumption (Power),
hours of computation (h), energy consumption (kWh $\cdot$ PUE)
	and carbon emissions (CO\textsubscript{2}e) of BERT
	vs.\ fastText. Training embeddings for all languages takes around 4 times the resources as training English. BERT numbers
	from \cite{strubell-etal-2019-energy}. We use our
	server's peak power consumption.
See appendix for details.\tablabel{emissions}}
\end{table}
\subsection{Vocabulary}
The vocabulary $\mathcal{V}$ of the wordpiece tokenizer is
of 
central importance for static embeddings as well as PLMs. BERT models come
with fixed vocabularies.
It would be prohibitive to
retrain the
models with a new vocabulary. It would also be too expensive to
increase the
vocabulary by a large factor: the embedding matrix is responsible for the
majority of the memory consumption of these models.

In contrast, increasing the vocabulary
size is cheap for static embeddings. We thus experiment with different vocabulary
sizes for static embeddings. To this end, we train new
vocabularies for each language on Wikipedia using the wordpiece
tokenizer \cite{schuster2012japanese}.

\subsection{Static Embeddings}

Using either newly trained vocabularies or existing
BERT vocabularies, we tokenize Wikipedia. We
then train fastText
embeddings \cite{bojanowski2017enriching} with default
parameters
(http://fasttext.cc).
We consider the same candidate set $\mathcal{C}$ as
for PLMs. Let 
$c \in \mathcal{C}$ be a candidate that gets split into
tokens $t_1, \dots, t_k$ by the wordpiece tokenizer. We then
assign to $c$ the embedding vector 
$$
\bar{e}_c = \frac{1}{k}\sum_{i = 1}^k e_{t_i}
$$ 
where $e_{t_i}$ is the fastText vector for token $t_i$. We
compute the representations for a query $q$ analogously.
For a query $q$ (the subject of a triple), we then compute
the prediction as:
$$
\arg\max_{c \in \mathcal{C}} \text{cosine-sim}(\bar{e}_q, \bar{e}_c), 
$$
i.e., we perform simple nearest neighbor matching. Note that
the static embedding method does not get any signal about
the relation. The method's only input is
the subject of a triple, and we leave incorporating a relation vector to future work.

\subsection{Evaluation Metric}
We compute precision at one  for each relation, i.e.,
$1/|T| \sum_{t \in T}\mathds{1}\{\hat{t}_{object} =
t_{object}\}$ where $T$ is the set of all triples and
$\hat{t}_{object}$
the object predicted using contextualized/static embeddings.
Note that $T$ is different for each language.
Our final measure (p1) is then
the 
precision at one (macro-)averaged over  relations.
As a consistency check we provide an \textbf{Oracle} baseline: it always predicts the most frequent object across triples based on the gold candidate sets.

\section{Results and Discussion}

In this section, we compare the performance of BERT and
fastText, 
analyze their resource consumption, 
and give evidence that
BERT composes meaningful representations from
subwords.

\subsection{BERT vs. fastText}
Results for English are in \tabref{english}. The table shows that when increasing the vocabulary
size, static embeddings and BERT exhibit similar performance on LAMA. The Oracle baseline is mostly outperformed. Only for small vocabulary sizes, fastText is worse. Performance of fastText increases with larger vocabulary sizes and with a
vocabulary size of 1000k we observe a 1.6\% absolute
performance increase of fastText embeddings compared to BERT
(41.2\% vs.\ 39.6\%). The performance gap between fastText
and BERT increases to 2.7\% points on LAMA-UHN, indicating
that fastText is less vulnerable to misleading
clues about the subject. 

Only providing results on English can be prone to unexpected
biases. Thus, we verify our results for nine additional
languages. Results are shown in \tabref{multilingual} and
the conclusions are similar: for large enough vocabularies,
static embeddings consistently have better performance. For languages outside the Indo-European
family, the performance gap between mBERT and fastText is much larger (e.g., 31.7 vs.\ 17.2 for Arabic) and mBERT is sometimes worse than the Oracle. 

Our fastText method is quite primitive: it is a 
type-restricted search for entities similar to what is most
prominent in the context (whose central element is the query
entity, e.g., ``Paris'' in (Ex1)). The fact that
fastText outperforms BERT raises the question: Does
BERT simply use associations between entities (like fastText)
or has it captured factual knowledge beyond this?

\subsection{BERT vs fastText: Diversity of Predictions}
The entropy of the distribution of predicted objects
is 6.5 for BERT vs.\ 7.3 for
fastText. So BERT's predictions are
less diverse.
Of 151 possible objects on average, BERT predicts
(on average) 85,
fastText 119.
For a given relation, BERT's  prediction tend to be dominated
by one object,
which is often the most frequent correct
object -- possibly because these objects are frequent in Wikipedia/Wikidata.  When filtering out triples whose
correct answer is the most frequent object,
BERT's performance drops to 35.7 whereas fastText's 
increases to 42.5.
See
\tabref{diversity} in the appendix for full results on diversity.
We leave
investigating why BERT has these narrower object preferences
for future work.

\begin{figure}
    \centering
     \includegraphics[width=1\linewidth]{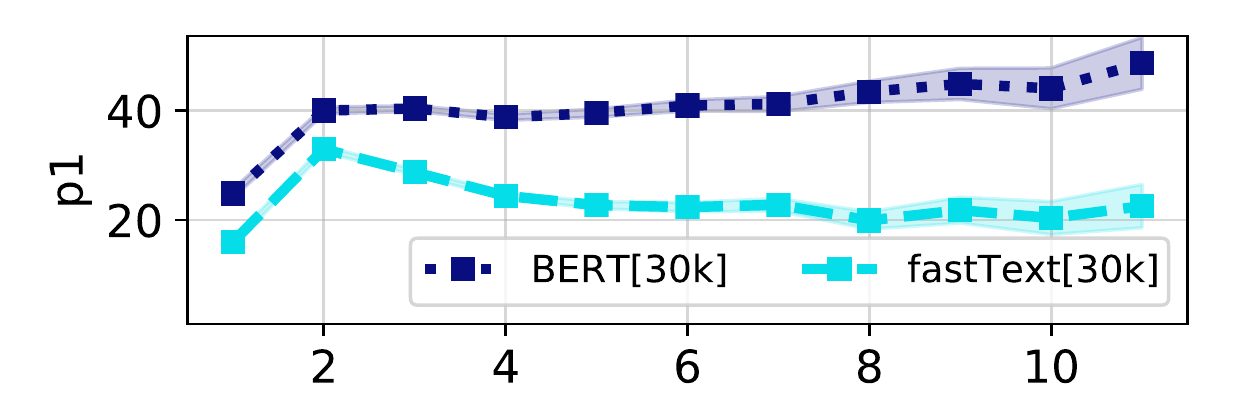}
     \includegraphics[width=1\linewidth]{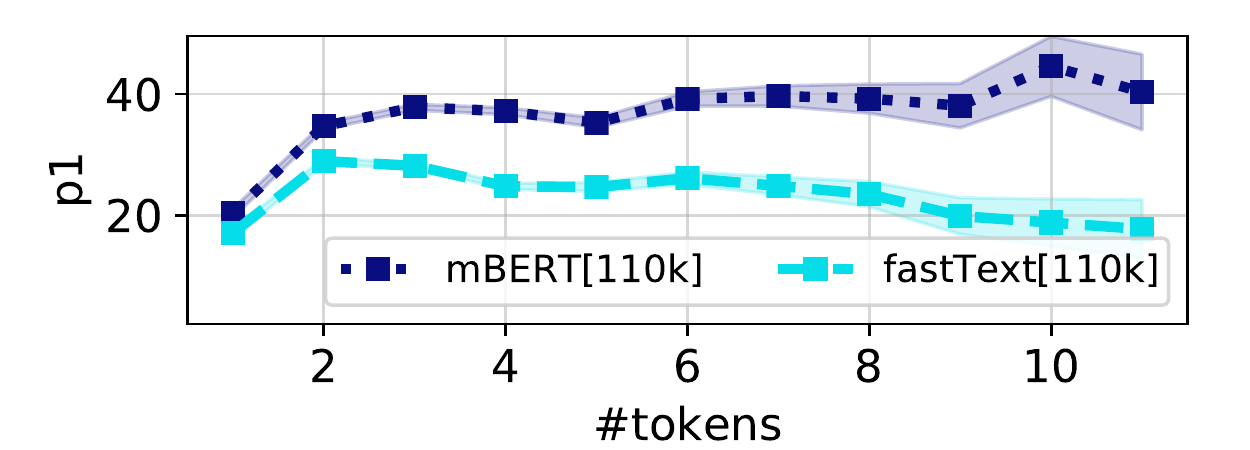}
    \caption{p1 as a function of the tokenization length of
    the triples' subjects. BERT and fastText use the same
    vocabulary here, ensuring comparability. BERT based models exhibit a stable performance independent of the number of tokens a subject gets split into. In contrast, fastText's performance drops.
    }
    \figlabel{length}
\end{figure}

\subsection{Contextualization in BERT}

BERT's attention mechanism should be able to handle  long
subjects -- 
in contrast to fastText, for which we  use simple
averaging. \figref{length} shows 
that fastText's performance indeed drops when the query gets tokenized into multiple tokens. In contrast, BERT's performance remains stable. We conclude that token
averaging harms fastText's performance and that the
attention mechanism in BERT composes meaningful representations from subwords.

We try to induce static embeddings from BERT by feeding
object and subject surface forms to BERT without any context
and then averaging the hidden representations for each
layer. \figref{layers}  analyzes whether a nearest neighbor
matching over this static embedding space extracted from
BERT's representations
is effective in extracting knowledge from it. We find that
performance on LAMA is significantly lower across all hidden
layers with the first two layers performing best.
That simple averaging does not work as well as
contextualization indicates that BERT is great at composing meaningful representations through attention. In future work, it would be interesting to extract better static representations from BERT, for example by extracting the representations of entities in real sentences.

\subsection{Resource Consumption}
\tabref{emissions} compares resource consumption of BERT vs.\ fastText
following
\citet{strubell-etal-2019-energy}. fastText can be efficiently computed on CPUs with a drastically lower power consumption and
computation time. Overall, fastText has only
0.3\% of the carbon emissions compared to BERT. In a recent study, \citet{zhang2020need} showed that capturing factual knowledge inside PLMs is an especially resource hungry task.

These big differences demonstrate that fastText, in
addition to performing better than BERT, is the
environmentally better model to ``encode knowledge'' of Wikipedia in an unsupervised fashion.
This calls into question the use of large  PLMs
as knowledge bases, particularly in light of 
the recent surge of knowledge augmented LMs, e.g., \cite{lewis2020pre,guu2020realm}.

\begin{figure}
	\centering
	\includegraphics[width=1\linewidth]{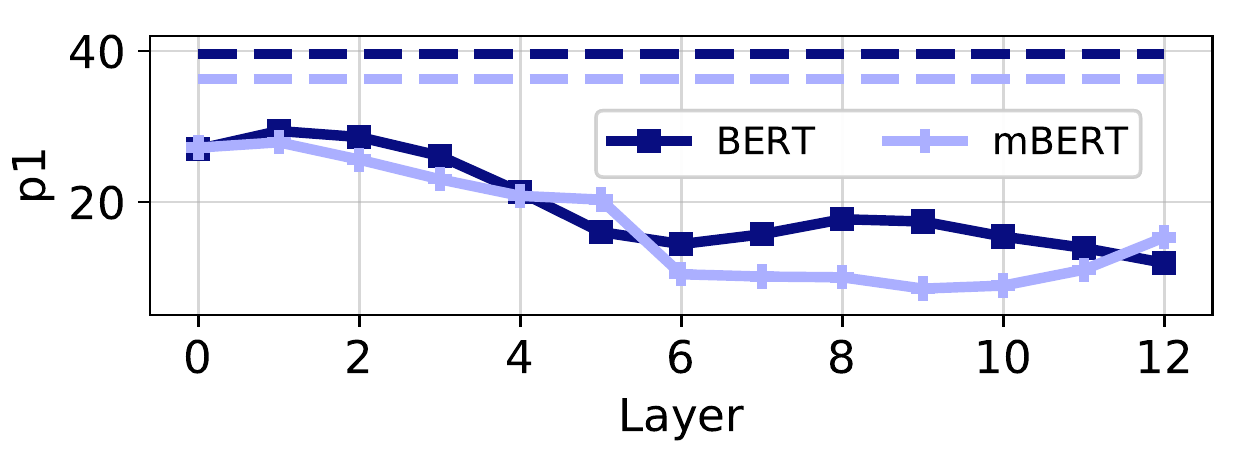}
	\caption{Contextualization in BERT. The dashed lines are p1 when querying with templates like ``Paris is the capital of [MASK].'' and a candidate set. The solid lines reflect performance of nearest neighbor matching with cosine similarity when inducing a static embedding space from the representations at these layers. This shows that extracting high quality static embeddings is not trivial, and BERT's contextualization is essential for getting  good performance.
	}
	\figlabel{layers}
\end{figure}

\section{Related Work}

\citet{petroni2019language} first asked: can
 PLMs function as KBs?
Subsequent
analysis focused on different aspects, such as negation
\cite{kassner-schutze-2020-negated, TACL1852}, easy to guess names
\cite{poerner-etal-2020-e}, finding alternatives to a
cloze-style approach
\cite{bouraoui2019inducing,heinzerling2020language,jiang2020can} or analyzing different  model sizes \cite{roberts2020knowledge}.

There is a recent surge of work that tries to improve  PLMs' ability to harvest factual knowledge:
\citet{zhang2019ernie}, \citet{Peters2019KnowledgeEC} and \citet{wang2020k} inject factual knowledge into  PLMs.
\citet{guu2020realm}, \citet{lewis2020pre}, \citet{Izacard2020LeveragingPR},
\citet{kassner2020bert} and \citet{petroni2020context} combine  PLMs  with information
retrieval
and \citet{Bosselut2019COMETCT}, \citet{weijie2019kbert} and \citet{Yu2020JAKETJP} with knowledge bases.

In contrast, we provide evidence that BERT's ability to answer factual queries 
is not more effective than capturing
``knowledge'' with simple traditional
static embeddings.  This suggests that
learning associations between entities and
type-restricted
similarity search over these associations may be at the core
of BERT's ability to answer
cloze-style KB queries, a new insight into BERT's working
mechanism.

\section{Conclusion}
We have shown that, when restricting cloze-style questions to a candidate set, static word embeddings 
outperform BERT. 
 To explain this puzzling superiority of a much simpler model,
we put forward a new characterization of factual
	knowledge learned by BERT: BERT seems to be able to
	complete cloze-style queries based on
	similarity assessments on a type-restricted vocabulary much like a nearest neighbor search for static embeddings.        

However, BERT may still be the better model for the task:
we assume perfect typing (for BERT and fastText) and
only evaluate  ranking. Typing is much
harder with static embeddings and BERT has been shown to
perform well at guessing the expected entity type based on a
template. BERT also works well with small vocabularies,
storing most of its ``knowledge'' in the
parameterization of subword composition.
	Our results suggest that increasing the vocabulary
        size and computing many more atomic entity
        representations with fastText is a cheap and
        environmentally friendly method of storing knowledge.
In contrast, learning high quality composition of smaller
units requires many more resources.

fastText is a simple cheap baseline that outperforms BERT on
LAMA, but was not considered in the original research.
This may be an example of a general problem:  ``green''
	baselines are often ignored, but should be  considered when evaluating
	resource-hungry deep learning models. A promising
        way forward would be to combine the best of both
        worlds, e.g., by building on 
in work that
incorporates large vocabularies into PLMs after pretraining.

\textbf{Acknowledgements.}
This work was supported
by the European Research Council (\# 740516) and the German
Federal Ministry of Education and Research (BMBF) under Grant
No. 01IS18036A. The authors of this work take full responsibility for its content. 
The first author was supported by the Bavarian research institute for digital transformation (bidt) through their fellowship program. 
We thank Yanai Elazar and the anonymous reviewers for valuable comments.

\bibliography{anthology,eacl2021}
\bibliographystyle{acl_natbib}
\appendix
\clearpage
\section{Resource Consumption}

We follow \citet{strubell-etal-2019-energy} for our computation. The measured peak energy consumption of our CPU-server was 618W. Considering the power usage effectiveness the required kWh are given by $p_t = 1.58\cdot t \cdot 618 / 1000$. Training the English fastText on Wikipedia took around 5 hours. Training all languages took 20 hours. The estimated CO\textsubscript{2}e can then be computed by $\text{CO\textsubscript{2}e} = 0.954\cdot p_t$

\section{Reproducibility Information}

For computation we use a CPU server with 96 CPU cores (Intel(R) Xeon(R) Platinum 8160) and 1024GB RAM. For BERT and mBERT inference we use a single GeForce GTX 1080Ti GPU.

Getting the object predictions for BERT and fastText is fast and takes a negligible amount of time. Training fastText embeddings takes between 1 to 5 hours depending on Wikipedia size.

BERT has around 110M parameters, mBERT around 178M. The fastText embeddings have $\mathcal{O}(nd)$ parameters where $n$ is the vocabulary size and $d$ is the embedding dimension. We use $d=300$. Thus, for most vocabulary sizes, fastText has significantly more parameters than the BERT models. But overall they are cheaper to train.

We did not perform any hyperparameter tuning. \tabref{hyperparams} gives an overview on third party software.
\tabref{ntriples} gives an overview on the number of triples in the dataset. Note that no training set is required, as all methods are completely unsupervised.

\begin{table}[b]
	\centering
	\footnotesize
	\begin{tabular}{crr}
		Language & \#Triples & \#Triples UHN \\
		\midrule
		\midrule
ar & 17129 & 13699\\
de & 29354 & 23493\\
en & 33981 & 27060\\
es & 28169 & 22683\\
fr & 30643 & 24487\\
he & 14769 & 12033\\
ja & 22920 & 17832\\
ko & 14217 & 11439\\
th & 8327 & 7065\\
tr & 13993 & 11274
	\end{tabular}
\caption{Overview on number of triples.\tablabel{ntriples}}
\end{table}

\begin{table}[t]
	\scriptsize
	\centering
	\begin{tabular}{lll}
		\textbf{System} & \textbf{Parameter} & \textbf{Value} \\
		\midrule
		\midrule
		\multirow{2}{1cm}{fastText}
		&Facebook Research & Version0.9.1 \\
		&Embedding Dimension&300\\
		\multirow{1}{1cm}{BERT}
		&Huggingface Transformer& Version 2.8.0 \\
		Tokenizers & Huggingface Tokenizers & Version 0.5.2\\
	\end{tabular}
	\caption{Overview on third party software.}
	\tablabel{hyperparams}
\end{table}

\section{Examples}

\tabref{examples} shows randomly sampled triples to perform an error analysis.

\section{Additional Results}

In this section we show additional results. \tabref{englishp5} shows the same as \tabref{english} but with precision at five. Analogously \tabref{multilingualp5}. \tabref{multilingualp1uhn} shows the same as \tabref{multilingual} but for LAMA-UHN. The trends and key insights are unchanged. 
\tabref{diversity} analyses the diversity of predictions by the different models.

\begin{table}
	\centering
	\footnotesize
	\def\mysep{0.08cm}
	\begin{tabular}{
			@{\hspace{\mysep}}l@{\hspace{\mysep}}
			@{\hspace{\mysep}}r@{\hspace{\mysep}}
			@{\hspace{\mysep}}r@{\hspace{\mysep}}
			@{\hspace{\mysep}}r@{\hspace{\mysep}}
			@{\hspace{\mysep}}r@{\hspace{\mysep}}
			@{\hspace{\mysep}}r@{\hspace{\mysep}}}
		\textbf{Model} & \textbf{Vocabulary Size} & \textbf{p1} & \textbf{p1-mf} & \textbf{entropy} &  \textbf{\#pred.}\\
		\midrule
		\midrule
		Oracle &  & 22.0 & 0.0 & 3.68 & 1 \\
\midrule BERT & 30k  & 39.6 & 35.7 & 6.48 & 85 \\
mBERT & 110k  & 36.3 & 32.6 & 6.41 & 86 \\
\midrule \multirow{7}{1cm}{fastText} & BERT-30k  & 26.9 & 27.7 & 7.04 & 107 \\
& mBERT-110k  & 27.5 & 27.6 & 7.09 & 110 \\
& 30k  & 16.4 & 15.9 & 7.13 & 111 \\
& 120k  & 34.3 & 35.4 & 7.30 & 115 \\
& 250k  & 37.7 & 38.9 & \textbf{7.33} & 118 \\
& 500k  & 39.9 & 41.2 & \textbf{7.33} & \textbf{119} \\
& 1000k  & \textbf{41.2} & \textbf{42.5} & 7.32 & \textbf{119} \\
	\end{tabular}
	\caption{Analysis of the diversity of predictions. \textit{p1-mf} is the p1 when excluding triples whose correct answer is the most frequent object. \textit{entropy} is the entropy of the distribution of predicted objects. \textit{\#pred.} denotes the average number of distinct objects predicted by the model across relations. The average number of unique objects in the candidate set across relations is 151. fastText has more diverse predictions, as the entropy is higher and the set of predicted objects is on average much larger.
		\tablabel{diversity}}
\end{table}

\begin{table}
	\centering
	\footnotesize
	\def\mysep{0.2cm}
	\begin{tabular}{
			@{\hspace{\mysep}}l@{\hspace{\mysep}}
			@{\hspace{\mysep}}r@{\hspace{\mysep}}
			@{\hspace{\mysep}}r@{\hspace{\mysep}}
			@{\hspace{\mysep}}r@{\hspace{\mysep}}}
		\textbf{Model} & \textbf{Vocabulary Size} & \multicolumn{2}{c}{\textbf{p5}}\\
		&&LAMA&LAMA-UHN\\
		\midrule
		\midrule
Oracle & & 48.0 & 49.7 \\
\midrule BERT & 30k & 64.1 & 57.9 \\
mBERT & 110k & 59.7 & 53.5 \\
\midrule \multirow{7}{1cm}{fastText} & BERT-30k & 48.7 & 41.9 \\
& mBERT-110k & 48.9 & 42.0 \\
& 30k & 26.3 & 16.5 \\
& 120k & 58.3 & 52.7 \\
& 250k & 62.7 & 58.1 \\
& 500k & 65.4 & 61.3 \\
& 1000k & \textbf{66.8 }& \textbf{63.1 }\\
	\end{tabular}
	\caption{Results for BERT, mBERT and
		fastText. Same as \tabref{english} but with p5.
		\tablabel{englishp5}}
\end{table}

\begin{table}
	\centering
	\scriptsize
	\def\mysep{0.05cm}
	\begin{tabular}{
			@{\hspace{\mysep}}l@{\hspace{\mysep}}
			@{\hspace{\mysep}}r@{\hspace{\mysep}}
			@{\hspace{\mysep}}r@{\hspace{\mysep}}
			@{\hspace{\mysep}}r@{\hspace{\mysep}}
			@{\hspace{\mysep}}r@{\hspace{\mysep}}
			@{\hspace{\mysep}}r@{\hspace{\mysep}}
			@{\hspace{\mysep}}r@{\hspace{\mysep}}
			@{\hspace{\mysep}}r@{\hspace{\mysep}}
			@{\hspace{\mysep}}r@{\hspace{\mysep}}
			@{\hspace{\mysep}}r@{\hspace{\mysep}}
			@{\hspace{\mysep}}r@{\hspace{\mysep}}}
		& \textbf{Vocab.}  & \multicolumn{9}{c}{\textbf{p5}}\\
		\textbf{Model} & \textbf{Size} & AR & DE & ES & FI & HE & JA & KO & TH & TR\\
		\midrule
		\midrule
Oracle & & 48.8 & 48.4 & 48.6 & 49.6 & 50.1 & 49.0 & 49.2 & 51.9 & 50.3 \\
\midrule mBERT & 110k & 33.8 & 51.3 & 53.9 & 46.2 & 38.2 & 36.5 & 43.0 & 37.0 & 55.5 \\
\midrule \multirow{6}{1cm}{fastText} & mBERT-110k & 26.0 & 40.5 & 42.9 & 43.8 & 27.7 & 24.0 & 31.9 & 33.9 & 50.3 \\
& 30k & 38.5 & 28.8 & 29.8 & 33.9 & 38.9 & 26.4 & 34.1 & 45.8 & 42.7 \\
& 120k & 51.6 & 48.9 & 55.2 & 49.7 & 54.1 & 44.1 & 54.8 & 56.0 & 60.9 \\
& 250k & 55.0 & 56.0 & 59.1 & 55.4 & 58.1 & 49.2 & 59.2 & 59.5 & 63.9 \\
& 500k & \textbf{57.0 }& 59.1 & 61.5 & 58.0 & \textbf{59.2 }& 50.9 & \textbf{59.7 }& \textbf{61.0 }& \textbf{64.6 }\\
& 1000k & 56.4 & \textbf{60.7 }& \textbf{62.2 }& \textbf{59.1 }& 58.9 & \textbf{51.7 }& 57.5 & 57.2 & 63.7 \\
	\end{tabular}
	\caption{p5 for mBERT and
		fastText on mLAMA. Numbers across languages are not
		comparable as the number of triples varies. 
		\tablabel{multilingualp5}}
\end{table}

\begin{table}[t]
	\centering
	\scriptsize
	\def\mysep{0.06cm}
	\begin{tabular}{
			@{\hspace{\mysep}}l@{\hspace{\mysep}}
			@{\hspace{\mysep}}r@{\hspace{\mysep}}
			@{\hspace{\mysep}}r@{\hspace{\mysep}}
			@{\hspace{\mysep}}r@{\hspace{\mysep}}
			@{\hspace{\mysep}}r@{\hspace{\mysep}}
			@{\hspace{\mysep}}r@{\hspace{\mysep}}
			@{\hspace{\mysep}}r@{\hspace{\mysep}}
			@{\hspace{\mysep}}r@{\hspace{\mysep}}
			@{\hspace{\mysep}}r@{\hspace{\mysep}}
			@{\hspace{\mysep}}r@{\hspace{\mysep}}
			@{\hspace{\mysep}}r@{\hspace{\mysep}}}
		& \textbf{Vocab.}  & \multicolumn{9}{c}{\textbf{p1}}\\
		\textbf{Model} & \textbf{Size} & AR & DE & ES & FI & HE & JA & KO & TH & TR\\
		\midrule
		\midrule
Oracle & & 23.1 & 23.8 & 23.2 & 22.9 & 24.5 & \textbf{22.5 }& 22.6 & 25.1 & 24.6 \\
\midrule mBERT & 110k & 12.1 & 26.1 & 27.6 & 15.8 & 11.0 & 11.8 & 15.1 & 10.8 & 27.7 \\
\midrule \multirow{6}{1cm}{fastText} & mBERT-110k & 7.8 & 14.3 & 16.9 & 15.0 & 6.6 & 6.4 & 8.0 & 7.4 & 19.4 \\
& 30k & 12.4 & 8.9 & 9.0 & 9.4 & 13.8 & 7.4 & 9.4 & 14.8 & 14.5 \\
& 120k & 20.2 & 18.9 & 23.8 & 18.1 & 22.1 & 15.4 & 21.0 & 23.8 & 26.1 \\
& 250k & 22.7 & 24.0 & 27.3 & 22.6 & 26.3 & 18.0 & 23.8 & \textbf{28.3 }& 28.7 \\
& 500k & \textbf{24.2 }& 26.6 & \textbf{30.1 }& 24.3 & 27.4 & 20.0 & \textbf{25.0 }& 27.6 & 29.4 \\
& 1000k & 23.7 & \textbf{27.6 }& \textbf{30.1 }& \textbf{25.6 }& \textbf{27.5 }& 20.4 & 23.2 & 27.2 & \textbf{29.8 }\\
	\end{tabular}
	\caption{p1 for mBERT and
		fastText on mLAMA-UHN. Numbers across languages are not
		comparable as the number of triples varies. 
		\tablabel{multilingualp1uhn}}
\end{table}

\begin{table*}[t]
	\tiny
	\centering
	\begin{tabular}{llllll}
		\textbf{Relation} & \textbf{Subject} & \textbf{Template} & \textbf{Object} & \textbf{BERT} & \textbf{fastText}\\
		\midrule
		\midrule
P1412 & William James & [X] used to communicate in [Y] . & English & English & Irish \\
P1412 & Bernardino Ochino & [X] used to communicate in [Y] . & Italian & Spanish & Italian \\
P1412 & Mick Lally & [X] used to communicate in [Y] . & Irish & English & Irish \\
P1412 & Robert Naunton & [X] used to communicate in [Y] . & English & English & Welsh \\
P108 & Steve Jobs & [X] works for [Y] . & Apple Inc. & Microsoft & Apple Inc. \\
P108 & Steve Wozniak & [X] works for [Y] . & Apple Inc. & CBS & Apple Inc. \\
P108 & Grady Booch & [X] works for [Y] . & IBM & IBM & Apple Inc. \\
P108 & Philip Don Estridge & [X] works for [Y] . & IBM & IBM & Apple Inc. \\
P178 & Safari & [X] is developed by [Y] . & Apple Inc. & Intel & Apple Inc. \\
P178 & PostScript & [X] is developed by [Y] . & Adobe & Microsoft & Adobe \\
P178 & Active Directory & [X] is developed by [Y] . & Microsoft & Microsoft & Apple Inc. \\
P178 & Internet Explorer & [X] is developed by [Y] . & Microsoft & Microsoft & Google \\
P31 & Long Preston & [X] is a [Y] . & village & village & pub \\
P31 & Israfil & [X] is a [Y] . & angel & village & angel \\
P31 & alfuzosin & [X] is a [Y] . & medication & protein & medication \\
P31 & Crawfordsburn & [X] is a [Y] . & village & village & suburb \\
P36 & Cook County & The capital of [X] is [Y] . & Chicago & Chicago & Williamson \\
P36 & Cayuga County & The capital of [X] is [Y] . & Auburn & Auburn & Greenville \\
P36 & Grand Est & The capital of [X] is [Y] . & Strasbourg & Paris & Strasbourg \\
P36 & Caddo Parish & The capital of [X] is [Y] . & Shreveport & Georgetown & Shreveport \\
P407 & The Vampyre & [X] was written in [Y] . & English & English & Gothic \\
P407 & Empire & [X] was written in [Y] . & English & English & Persian \\
P407 & Politika & [X] was written in [Y] . & Serbian & Latin & Serbian \\
P407 & Lenta.ru & [X] was written in [Y] . & Russian & German & Russian \\
P449 & Drake \& Josh & [X] was originally aired on [Y] . & Nickelodeon & Nickelodeon & Fox Arena \\
P449 & Salute Your Shorts & [X] was originally aired on [Y] . & Nickelodeon & Nickelodeon & Lifetime \\
P449 & Yo Momma & [X] was originally aired on [Y] . & MTV & CBS & MTV \\
P449 & Hey Arnold! & [X] was originally aired on [Y] . & Nickelodeon & CBS & Nickelodeon \\
P127 & Xbox & [X] is owned by [Y] . & Microsoft & Microsoft & Nintendo \\
P127 & Eiffel Tower & [X] is owned by [Y] . & Paris & Boeing & Paris \\
P127 & Lotus Software & [X] is owned by [Y] . & IBM & IBM & Microsoft \\
P127 & Lexus & [X] is owned by [Y] . & Toyota & Chrysler & Toyota \\
P364 & Black Narcissus & The original language of [X] is [Y] . & English & English & Irish \\
P364 & The God Delusion & The original language of [X] is [Y] . & English & English & Hebrew \\
P364 & Vecinos & The original language of [X] is [Y] . & Spanish & Latin & Spanish \\
P364 & Janji Joni & The original language of [X] is [Y] . & Indonesian & Marathi & Indonesian \\
P106 & Halle Berry & [X] is a [Y] by profession . & model & model & organist \\
P106 & Gregory Chamitoff & [X] is a [Y] by profession . & astronaut & lawyer & astronaut \\
P106 & Karl Taylor Compton & [X] is a [Y] by profession . & physicist & lawyer & physicist \\
P106 & Herbert Romulus O'Conor & [X] is a [Y] by profession . & lawyer & lawyer & playwright \\
P176 & System Controller Hub & [X] is produced by [Y] . & Intel & Intel & Apple Inc. \\
P176 & Daihatsu Boon & [X] is produced by [Y] . & Toyota & Honda & Toyota \\
P176 & British Rail Class 360 & [X] is produced by [Y] . & Siemens & Siemens & Volvo Cars \\
P176 & Dino & [X] is produced by [Y] . & Ferrari & Sony & Ferrari \\
P937 & Howard Florey & [X] used to work in [Y] . & London & London & Montgomery \\
P937 & Alberts Kviesis & [X] used to work in [Y] . & Riga & Stockholm & Riga \\
P937 & Ramsay MacDonald & [X] used to work in [Y] . & London & London & Scotland \\
P937 & Juan March & [X] used to work in [Y] . & Madrid & Paris & Madrid \\
P463 & United States of America & [X] is a member of [Y] . & NATO & NATO & PBS \\
P463 & Croatia & [X] is a member of [Y] . & NATO & NATO & FIFA \\
P463 & Mexico national football team & [X] is a member of [Y] . & FIFA & CONCACAF & FIFA \\
P463 & Estonia & [X] is a member of [Y] . & NATO & FIFA & NATO \\
P138 & Germany & [X] is named after [Y] . & Bavaria & France & Bavaria \\
P138 & GNU & [X] is named after [Y] . & Unix & Aristotle & Unix \\
P138 & solar mass & [X] is named after [Y] . & Sun & Sun & carbon \\
P138 & Torino F.C. & [X] is named after [Y] . & Turin & Turin & Apple Inc. \\
P101 & Edward Burnett Tylor & [X] works in the field of [Y] . & anthropology & medicine & anthropology \\
P101 & Anaxagoras & [X] works in the field of [Y] . & philosophy & philosophy & philosopher \\
P101 & Adam Carolla & [X] works in the field of [Y] . & comedian & psychology & comedian \\
P101 & physical system & [X] works in the field of [Y] . & physics & physics & physiology \\
P39 & Augustine Kandathil & [X] has the position of [Y] . & archbishop & minister & archbishop \\
P39 & John XXI & [X] has the position of [Y] . & pope & bishop & pope \\
P39 & Photinus of Sirmium & [X] has the position of [Y] . & bishop & bishop & pope \\
P39 & Samson of Dol & [X] has the position of [Y] . & bishop & bishop & God \\
P530 & Holy See & [X] maintains diplomatic relations with [Y] . & Italy & Italy & Austria \\
P530 & Malta & [X] maintains diplomatic relations with [Y] . & Italy & Italy & Malta \\
P530 & Liechtenstein & [X] maintains diplomatic relations with [Y] . & Austria & Switzerland & Austria \\
P530 & Saudi Arabia & [X] maintains diplomatic relations with [Y] . & Kuwait & Qatar & Kuwait \\
P264 & Georg Solti & [X] is represented by music label [Y] . & Decca & EMI & Decca \\
P264 & The Temptations & [X] is represented by music label [Y] . & Motown & EMI & Motown \\
P264 & David Bowie & [X] is represented by music label [Y] . & EMI & EMI & Barclay \\
P264 & Maria Callas & [X] is represented by music label [Y] . & EMI & EMI & Decca \\
P1376 & Florence & [X] is the capital of [Y] . & Tuscany & Italy & Tuscany \\
P1376 & Canberra & [X] is the capital of [Y] . & Australia & Australia & Queensland \\
P1376 & Heraklion & [X] is the capital of [Y] . & Crete & Greece & Crete \\
P1376 & Islamabad & [X] is the capital of [Y] . & Pakistan & Pakistan & Karachi \\
P1001 & Jatiya Sangshad & [X] is a legal term in [Y] . & Bangladesh & India & Bangladesh \\
P1001 & Legislative Yuan & [X] is a legal term in [Y] . & Taiwan & Singapore & Taiwan \\
P1001 & Manitoba Act, 1870 & [X] is a legal term in [Y] . & Canada & Canada & Ontario \\
P1001 & Yang di-Pertuan Agong & [X] is a legal term in [Y] . & Malaysia & Malaysia & Brunei \\
P495 & soppressata & [X] was created in [Y] . & Italy & Italy & Peru \\
P495 & Kefalotyri & [X] was created in [Y] . & Greece & Cyprus & Greece \\
P495 & Degrassi High & [X] was created in [Y] . & Canada & Canada & Jordan \\
P495 & Fox Soccer News & [X] was created in [Y] . & Canada & Australia & Canada \\
	\end{tabular}
	\caption{We sample two random triples where either BERT or fastText[1000k] is correct per relation. One can see for example that BERT mostly predicts ``jazz'' for relation P136.}
	\tablabel{examples}
\end{table*}

\begin{table*}[t]
	\tiny
	\centering
	\begin{tabular}{llllll}
		\textbf{Relation} & \textbf{Subject} & \textbf{Template} & \textbf{Object} & \textbf{BERT} & \textbf{fastText}\\
		\midrule
		\midrule
P527 & army & [X] consists of [Y] . & infantry & infantry & cavalry \\
P527 & Windward Islands & [X] consists of [Y] . & Barbados & Bermuda & Barbados \\
P527 & taxon & [X] consists of [Y] . & organism & grass & organism \\
P527 & humanities & [X] consists of [Y] . & art & art & linguistics \\
P1303 & Kenny G & [X] plays [Y] . & saxophone & guitar & saxophone \\
P1303 & Stuart Duncan & [X] plays [Y] . & fiddle & guitar & fiddle \\
P1303 & Herbie Nichols & [X] plays [Y] . & piano & piano & harmonica \\
P1303 & Nat King Cole & [X] plays [Y] . & piano & piano & saxophone \\
P190 & Uzhhorod & [X] and [Y] are twin cities . & Moscow & Moscow & Lviv \\
P190 & Vienna & [X] and [Y] are twin cities . & Budapest & Budapest & Vienna \\
P190 & Cali & [X] and [Y] are twin cities . & Guadalajara & Santiago & Guadalajara \\
P190 & Mindelo & [X] and [Y] are twin cities . & Porto & Santiago & Porto \\
P47 & Monreale & [X] shares border with [Y] . & Palermo & Italy & Palermo \\
P47 & Afghanistan & [X] shares border with [Y] . & Pakistan & Pakistan & Afghanistan \\
P47 & Ukraine & [X] shares border with [Y] . & Russia & Russia & Ukraine \\
P47 & Edegem & [X] shares border with [Y] . & Antwerp & Ethiopia & Antwerp \\
P30 & McDonald Heights & [X] is located in [Y] . & Antarctica & Africa & Antarctica \\
P30 & Balham Valley & [X] is located in [Y] . & Antarctica & Antarctica & Africa \\
P30 & Southern Netherlands & [X] is located in [Y] . & Europe & Europe & Africa \\
P30 & Pitcairn Islands & [X] is located in [Y] . & Oceania & Antarctica & Oceania \\
P361 & arithmetic & [X] is part of [Y] . & mathematics & mathematics & logic \\
P361 & agricultural science & [X] is part of [Y] . & agriculture & agriculture & science \\
P361 & zoology & [X] is part of [Y] . & biology & science & biology \\
P361 & neuroscience & [X] is part of [Y] . & psychology & science & psychology \\
P103 & Muppalaneni Shiva & The native language of [X] is [Y] . & Telugu & Marathi & Telugu \\
P103 & Joseph Reinach & The native language of [X] is [Y] . & French & English & French \\
P103 & Raymond Queneau & The native language of [X] is [Y] . & French & French & Breton \\
P103 & Lindsey Davis & The native language of [X] is [Y] . & English & English & Welsh \\
P20 & James Northcote & [X] died in [Y] . & London & London & Morris \\
P20 & George Frampton & [X] died in [Y] . & London & London & Chapman \\
P20 & Peter Strudel & [X] died in [Y] . & Vienna & Paris & Vienna \\
P20 & Gaetano Gandolfi & [X] died in [Y] . & Bologna & Rome & Bologna \\
P27 & August Gailit & [X] is [Y] citizen . & Estonia & Luxembourg & Estonia \\
P27 & Ada Yonath & [X] is [Y] citizen . & Israel & India & Israel \\
P27 & Enrique Llanes & [X] is [Y] citizen . & Mexico & Mexico & Spain \\
P27 & Timothy Anglin & [X] is [Y] citizen . & Canada & Canada & England \\
P279 & Ciliary neurotrophic factor & [X] is a subclass of [Y] . & protein & protein & inflammation \\
P279 & Decorin & [X] is a subclass of [Y] . & protein & protein & perfume \\
P279 & shinto shrine & [X] is a subclass of [Y] . & sanctuary & Buddhism & sanctuary \\
P279 & articled clerk & [X] is a subclass of [Y] . & apprentice & jurist & apprentice \\
P19 & Frans Floris I & [X] was born in [Y] . & Antwerp & Amsterdam & Antwerp \\
P19 & Sajjad Ali & [X] was born in [Y] . & Lahore & Tehran & Lahore \\
P19 & Henry Mayhew & [X] was born in [Y] . & London & London & Fowler \\
P19 & Rob Lee & [X] was born in [Y] . & London & London & Gary \\
P159 & Swedish Orphan Biovitrum & The headquarter of [X] is in [Y] . & Stockholm & Stockholm & Gothenburg \\
P159 & Canadian Jewish Congress & The headquarter of [X] is in [Y] . & Ottawa & Ottawa & Winnipeg \\
P159 & Florida International University & The headquarter of [X] is in [Y] . & Miami & Tampa & Miami \\
P159 & Edipresse & The headquarter of [X] is in [Y] . & Lausanne & Chennai & Lausanne \\
P413 & Markus Halsti & [X] plays in [Y] position . & midfielder & midfielder & goaltender \\
P413 & Luca Danilo Fusi & [X] plays in [Y] position . & midfielder & midfielder & goalkeeper \\
P413 & Mike Teel & [X] plays in [Y] position . & quarterback & forward & quarterback \\
P413 & Doug Buffone & [X] plays in [Y] position . & linebacker & forward & linebacker \\
P37 & Sorengo & The official language of [X] is [Y] . & Italian & Portuguese & Italian \\
P37 & Padasjoki & The official language of [X] is [Y] . & Finnish & English & Finnish \\
P37 & Wallonia & The official language of [X] is [Y] . & French & French & Basque \\
P37 & Biel/Bienne & The official language of [X] is [Y] . & French & French & Czech \\
P140 & Gautama Buddha & [X] is affiliated with the [Y] religion . & Buddhism & Hindu & Buddhism \\
P140 & Christianization & [X] is affiliated with the [Y] religion . & Christianity & Christian & Christianity \\
P140 & Albanians & [X] is affiliated with the [Y] religion . & Christian & Christian & Muslim \\
P740 & SNCF & [X] was founded in [Y] . & Paris & Paris & France \\
P740 & Odex & [X] was founded in [Y] . & Singapore & Germany & Singapore \\
P740 & Comerica & [X] was founded in [Y] . & Detroit & Prague & Detroit \\
P740 & Pink Fairies & [X] was founded in [Y] . & London & London & Gold \\
P276 & Saint-Domingue expedition & [X] is located in [Y] . & Haiti & France & Haiti \\
P276 & 2002 Australian Op[X] is located in [Y] . & Melbourne & Melbourne & Australia \\
P276 & 2013 German federal election & [X] is located in [Y] . & Germany & Berlin & Germany \\
P276 & Cantabrian Wars & [X] is located in [Y] . & Spain & Spain & Catalonia \\
P136 & Giulio Caccini & [X] plays [Y] music . & opera & jazz & opera \\
P136 & Nicolas Dalayrac & [X] plays [Y] music . & opera & jazz & opera \\
P136 & Georgie Auld & [X] plays [Y] music . & jazz & jazz & ballad \\
P136 & Chess Records & [X] plays [Y] music . & jazz & jazz & reggae \\
P17 & Eibenstock & [X] is located in [Y] . & Germany & Germany & Austria \\
P17 & Vrienden van het Platteland & [X] is located in [Y] . & Netherlands & Belgium & Netherlands \\
P17 & Fawkner & [X] is located in [Y] . & Australia & Lebanon & Australia \\
P17 & Wakefield Park & [X] is located in [Y] . & Australia & Australia & The Bahamas \\
P131 & Squantz Pond State Park & [X] is located in [Y] . & Connecticut & Somerset & Connecticut \\
P131 & Ballyfermot & [X] is located in [Y] . & Dublin & Ireland & Dublin \\
P131 & Downtown East Village, Calgary & [X] is located in [Y] . & Alberta & Alberta & Toronto \\
P131 & Edmonton City Centre Airport & [X] is located in [Y] . & Alberta & Alberta & Toronto \\
	\end{tabular}
\caption{\tabref{examples} continued.}
\end{table*}

\end{document}